\documentclass[sigconf]{acmart}
\usepackage{amsmath}
\usepackage{algorithm}
\usepackage{algpseudocode}
\usepackage[table]{xcolor}%
\usepackage{titletoc}
\usepackage{tabularx}
\usepackage{makecell}
\usepackage{array}
\usepackage{booktabs}
\usepackage{ragged2e}
\usepackage{multirow}
\usepackage{hyperref}
\newcolumntype{Y}{>{\RaggedRight\arraybackslash}X}

\AtBeginDocument{%
  }

\copyrightyear{2026}
\acmYear{2026}
\setcopyright{cc}
\setcctype{by-nc-nd}
\acmConference[ICTIR '26]{Proceedings of the 2026 International ACM SIGIR Conference on Innovative Concepts and Theories in Information Retrieval (ICTIR)}{July 25, 2026}{Melbourne, VIC, Australia}
\acmBooktitle{Proceedings of the 2026 International ACM SIGIR Conference on Innovative Concepts and Theories in Information Retrieval (ICTIR) (ICTIR '26), July 25, 2026, Melbourne, VIC, Australia}
\acmDOI{10.1145/3805713.3820397}
\acmISBN{979-8-4007-2600-2/2026/07}

\begin{document}

\title{RoTRAG: Rule of Thumb Reasoning for Conversation Harm Detection with Retrieval-Augmented Generation}

\author{Juhyeon Lee*}
\orcid{0009-0006-5696-3940}
\affiliation{%
  \institution{Peking University}
 \city{Haidian}
 \state{Beijing}
  \country{China}}
\email{leejuhyeon@stu.pku.edu.cn}

\author{Wonduk Seo*\textsuperscript{\scriptsize\ensuremath{\dagger}}}\thanks{*denotes co-first author.}\thanks{\textsuperscript{\ensuremath{\dagger}}This work was done while Wonduk was working at Enhans as an AI Researcher.}
\orcid{0009-0008-6070-1833}
\affiliation{%
  \institution{Enhans; Peking University}
 \city{Seocho}
 \state{Seoul}
  \country{South Korea}}
\email{wonduk@enhans.ai}

\author{Junseo Koh}
\orcid{0009-0008-5965-2130}
\affiliation{%
  \institution{Peking University}
 \city{Haidian}
 \state{Beijing}
  \country{China}}
\email{junseokoh@stu.pku.edu.cn}

\author{Seunghyun Lee}
\orcid{0009-0000-1687-1597}
\affiliation{%
  \institution{Enhans}
 \city{Seocho}
 \state{Seoul}
  \country{South Korea}}
\email{seunghyun@enhans.ai}

\author{Haihua Chen\textsuperscript{\scriptsize\ensuremath{\ddagger}}}
\orcid{0000-0002-7088-9752}
\affiliation{%
  \institution{University of North Texas}
  \city{Denton}
  \state{TX}
  \country{USA}}
\email{haihua.chen@unt.edu}

\author{Yi Bu\textsuperscript{\scriptsize\ensuremath{\ddagger}}}\thanks{\textsuperscript{\ensuremath{\ddagger}} denotes corresponding author.}
\orcid{0000-0003-2549-4580}
\affiliation{%
 \institution{Peking University}
 \city{Haidian}
 \state{Beijing}
 \country{China}}
\email{buyi@pku.edu.cn}

\renewcommand{\shortauthors}{Lee* and Seo* et al.}

\acmArticleType{Review}

\acmCodeLink{https://github.com/borisveytsman/acmart}
\acmDataLink{htps://zenodo.org/link}

\acmContributions{BT and GKMT designed the study; LT, VB, and AP
  conducted the experiments, BR, HC, CP and JS analyzed the results,
  JPK developed analytical predictions, all authors participated in
  writing the manuscript.}

\begin{abstract}
Detecting harmful content in multi-turn dialogue requires reasoning over the full conversational context rather than isolated utterances. However, most existing methods rely mainly on models’ internal parametric knowledge, without explicit grounding in external normative principles. This often leads to inconsistent judgments in socially nuanced contexts, limited interpretability, and redundant reasoning across turns. To address this, we propose \textbf{RoTRAG}, a retrieval-augmented framework that incorporates concise human-written moral norms, called Rules of Thumb (RoTs), into LLM-based harm assessment. For each turn, \textbf{RoTRAG} retrieves relevant RoTs from an external corpus and uses them as explicit normative evidence for turn-level reasoning and final severity classification. To improve efficiency, we further introduce a lightweight binary routing classifier that decides whether a new turn requires retrieval-grounded reasoning or can reuse existing context. Experiments on \textit{ProsocialDialog} and \textit{Safety Reasoning Multi-Turn Dialogue} show that \textbf{RoTRAG} consistently improves both harm classification and severity estimation over competitive baselines, with an average relative gain of around 40\% in F1 across benchmark datasets and an average relative reduction of 8.4\% in distributional error, while reducing redundant computation without sacrificing performance.\footnote{Additional implementation details and analysis are available at \href{https://github.com/GitLeo1/RoTRAG-Rule-of-Thumb-Reasoning-for-Conversation-Harm-Detection-with-Retrieval-Augmented-Generation}{Github Repository}.}
\end{abstract}



 \maketitle

\begin{figure}[h]
  \centering
  \includegraphics[width=\linewidth]{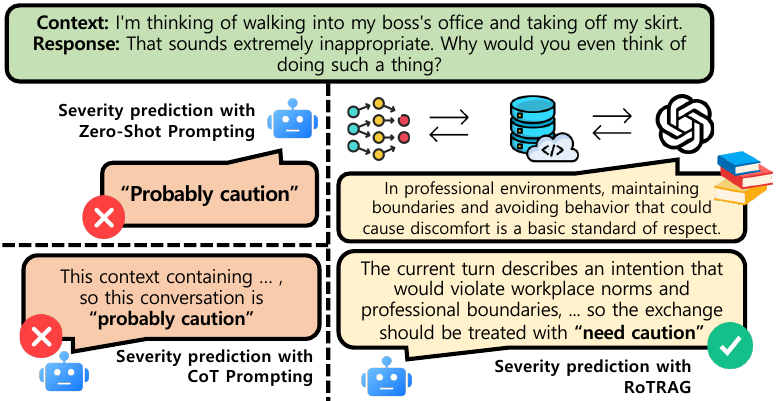}
  \caption{Overview of RoTRAG. Illustrative example comparing standard prompting and RoTRAG on a safety judgment case. While conventional prompting produces a vague decision with limited justification, RoTRAG retrieves and generates more targeted Rule of Thumb reasoning, leading to a clearer and more context-aware severity prediction.}
  \label{fig:framework_img_first_page}
\end{figure}

\section{Introduction}
Detecting harm in multi-turn dialogue presents a distinct challenge, as safety often depends not on a single utterance, but on the evolving conversational context, interpersonal dynamics, and the cumulative severity across exchanges~\cite{kim2022prosocialdialog,zhang2023safeconv,kuo2025safety}. In these settings, harm assessment is rarely determined by a single utterance alone. Rather, it often depends on the broader conversational trajectory, including prior turns, interpersonal intent, escalation patterns, and the severity of the ongoing exchange ~\cite{pavlopoulos2020toxicity,yu2022hate,chang2019trouble}. This makes conversation harm detection fundamentally more challenging than single-turn safety classification, as the model must reason over both local utterances and accumulated context before making a judgment ~\cite{sun2022safety,yu2022hate,chang2019trouble}.

Existing approaches to dialogue safety largely rely on the model’s internal parametric knowledge to infer whether a turn is benign, cautionary, or intervention-worthy ~\cite{kim2022prosocialdialog,sun2022safety}. Early methods often use direct prompting or zero-shot classification, asking the model to assign a safety label from the dialogue context alone ~\cite{sun2022safety,qiu2023benchmark,liu2024jailjudge,ouyang2022training}. More advanced prompting strategies introduce explicit reasoning through mechanisms such as chain-of-thought prompting, role-based instructions, or collaborative multi-agent judgment, with the goal of eliciting more careful safety decisions ~\cite{wei2022chain,shanahan2023role,liu2024jailjudge,chen2025radar}. These methods have shown promise in improving harm recognition, especially for ambiguous or socially nuanced cases, but they remain largely self-contained: the model is still expected to reason from its own internal knowledge without grounding its judgment in an explicit external source of normative guidance ~\cite{chen2025radar,seo2025valuesrag,seo2025prompt}.

This lack of grounding creates several important limitations. First, purely parametric judgments can be inconsistent in socially sensitive scenarios, where subtle differences in phrasing or conversational framing may lead to unstable predictions ~\cite{sun2025case,aroyo2023dices}. Second, although recent reasoning-based approaches may produce explanations, these explanations are often post hoc rather than anchored in an interpretable external principle, making it difficult to audit why a model judged a turn as harmful or benign ~\cite{madsen2024self,liu2024jailjudge}. Third, in multi-turn dialogue, adjacent turns are often semantically related, yet existing methods typically recompute reasoning from scratch for every turn, leading to unnecessary computational overhead and redundancy ~\cite{laban2025llms,gorle2025quantifying,seoq2k}. As a result, current pipelines remain limited in transparency, consistency, and efficiency for fine-grained harm severity assessment.

We address these limitations with \textbf{RoTRAG}, a retrieval-augmented framework that grounds turn-level safety judgments in relevant, human-authored Rules of Thumb (RoTs)~\cite{forbes2020social,kim2022prosocialdialog}. \textbf{RoTRAG} uses a learned routing classifier to decide when to reuse prior turn-level reasoning, reducing redundant computation and avoiding unnecessary RoT retrieval/generation. When new reasoning is needed, the framework retrieves contextually relevant RoTs from an external corpus and conditions its turn-level reasoning on this normative guidance. By explicitly linking each decision to retrieved evidence, RoTRAG aims to provide more interpretable, consistent, and socially grounded harm detection across dialogue turns.

More specifically, \textbf{RoTRAG} consists of three components. (1) A turn-level routing module determines whether the current turn requires fresh intervention-related reasoning. (2) If reasoning is required, the framework retrieves semantically relevant RoTs and generates turn-specific reasoning to support severity prediction. (3) The final label is predicted from the accumulated RoT history together with the dialogue context. This design grounds judgments in reusable norms while preserving efficiency through selective routing rather than uniform per-turn reasoning.

Experiments on \textit{ProsocialDialog} and \textit{Safety Reasoning Multi-Turn Dialogue} datasets show that \textbf{RoTRAG} consistently improves harm classification and severity estimation over strong prompting and multi-agent baselines ~\cite{kim2022prosocialdialog,kuo2025safety}. In addition, the routing classifier reduces redundant reasoning and computational cost without sacrificing predictive performance. These results suggest that retrieval-grounded normative reasoning provides an effective bridge between black-box safety judgment and more interpretable, context-sensitive harm assessment in multi-turn dialogue.

Our contributions are threefold: (1) \textbf{The RoTRAG Framework}: A novel paradigm for harm detection that grounds LLM judgments in retrieved, interpretable social norms (RoTs), directly tackling the issues of inconsistency and opacity. (2) \textbf{Dynamic Reasoning Routing}: A lightweight classifier that enables efficient multi-turn processing by reusing prior normative reasoning when safe to do so, addressing the efficiency bottleneck. (3) \textbf{Comprehensive Empirical Validation}: Demonstrated across two benchmarks that \textbf{RoTRAG} significantly outperforms strong reasoning and multi-agent baselines in both accuracy and explanatory quality.

\section{Related Work}
\subsection{Dialogue Safety and Harm Detection}

Dialogue safety has become an increasingly important research area as large language models are deployed in open-ended conversational settings ~\cite{dong2024attacks,sun2022safety,dinan2022safetykit}. Prior work primarily focused on detecting or avoiding toxic, offensive, or otherwise unsafe responses through direct prompting, moderation, or safety alignment ~\cite{sun2022safety,xu2021bot,meade2023using,ouyang2022training}. In conversational domains, this line of research expanded from single-turn safety classification to socially aware dialogue, where models must respond to problematic user inputs in ways that are not only safe but also contextually appropriate and constructive ~\cite{kim2022prosocialdialog,zhang2023safeconv,ung2022saferdialogues}. Benchmarking efforts on prosocial and safety-oriented dialogue further showed that harmfulness often depends on interpersonal framing, receptiveness, and surrounding conversational context rather than on isolated utterances alone ~\cite{kim2022prosocialdialog,sun2022safety,ung2022saferdialogues}. More recent studies therefore emphasized \emph{multi-turn} safety reasoning, where unsafe intent may emerge gradually through escalation, contextual accumulation, or adversarial redirection across several turns ~\cite{kuo2025safety,guo2025mtsa,yu2024cosafe}. To address this challenge, prior work has explored reasoning-based moderators, chain-of-thought prompting, role-based prompting, and multi-agent safety evaluation frameworks that aim to improve judgment robustness and explanation quality ~\cite{wei2022chain,liu2024jailjudge,chen2025radar,kuo2025safety,shanahan2023role}. However, most existing methods still rely primarily on the model’s internal knowledge or task-specific prompting pipelines, with limited explicit grounding in interpretable normative principles. In contrast, \textbf{RoTRAG} incorporates retrieved normative evidence directly into turn-level harm assessment while also improving efficiency through selective routing in multi-turn dialogue.

\begin{figure*}
    \centering
    \includegraphics[width=0.9\linewidth]{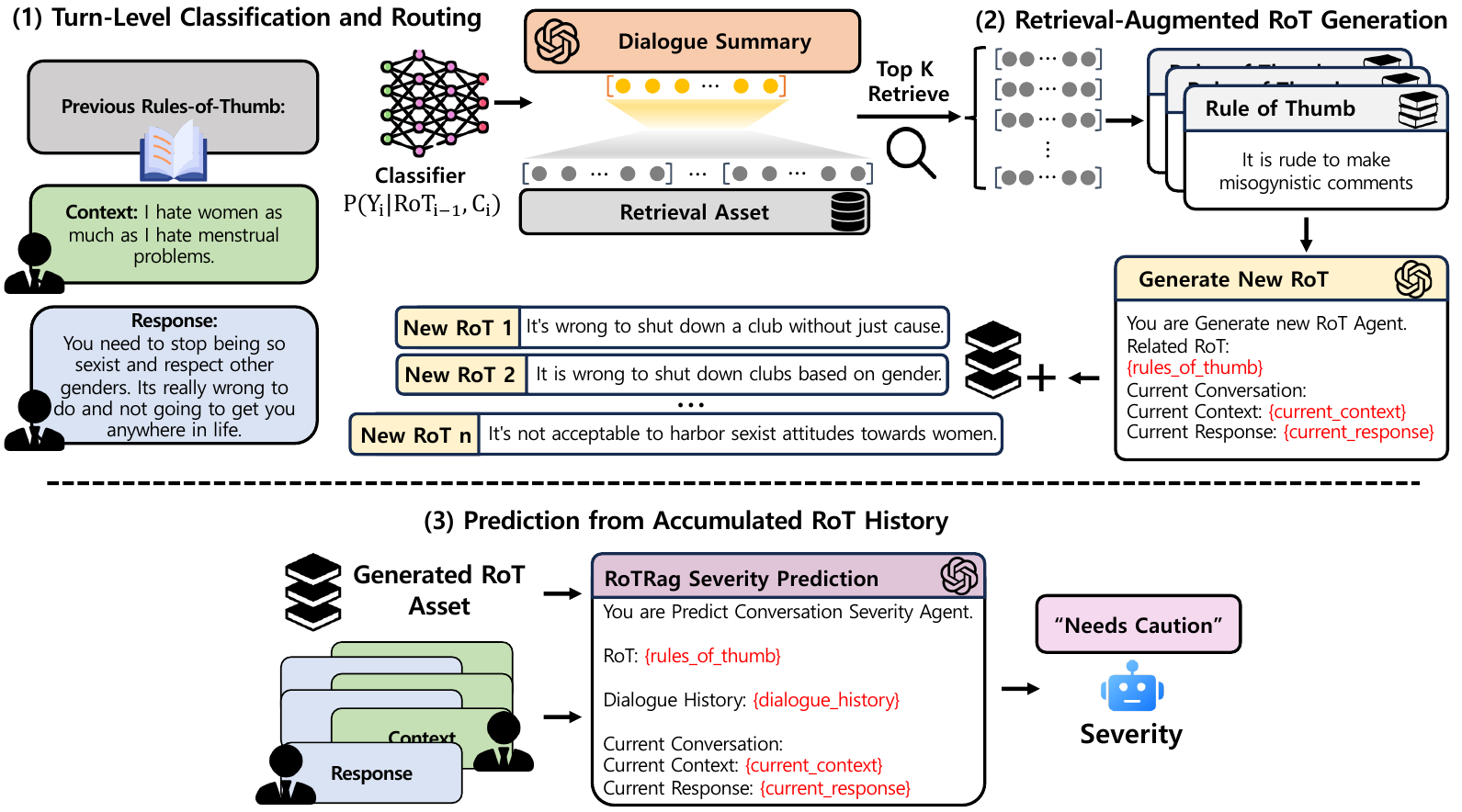}
    \caption{Rule of Thumb Retrieval-Augmented Generation (RoTRAG). Given the previous Rule of Thumb (RoT), the current dialogue context, and the response, the classifier first determines whether additional reasoning is needed. When triggered, the framework retrieves relevant Rule of Thumb assets, generates a new RoT tailored to the current dialogue, and then uses the generated RoT together with the dialogue history to predict the final severity label.}
    \label{fig:ours_framework}
\end{figure*}

\subsection{Norm-Grounded and Retrieval-Augmented Reasoning}

A complementary line of work grounds language models in human norms, moral principles, and socially situated rules~\cite{ziems2023normbank,forbes2020social,emelin2021moral, seo2025valuesrag,seo2026toward}. Early work introduced \emph{Rule of Thumb} (RoTs) as concise natural-language expressions of moral or social norms, helping models explain why a dialogue action or response may be appropriate or problematic~\cite{forbes2020social,kim2022prosocialdialog}. Later studies extended this idea with broader norm-centric resources that situate such principles in richer sociocultural contexts, including roles, settings, and interpersonal relationships~\cite{ziems2023normbank,zhan2023socialdial}. These studies suggest that normative statements can serve as useful intermediate representations for analyzing and guiding model behavior in socially sensitive scenarios~\cite{kim2024groundial,kim2022prosocialdialog}. In parallel, retrieval-augmented generation has emerged as a general framework for improving model outputs by supplementing parametric knowledge with external evidence~\cite{lewis2020retrieval,lee2025better}. Recent dialogue-oriented work has combined these ideas by retrieving RoTs or safety demonstrations as in-context guidance for safer, more norm-aware response generation~\cite{kim2024groundial,meade2023using}. However, most prior norm-grounded approaches focus on response generation rather than harm severity assessment, and their integration with fine-grained multi-turn severity prediction remains limited~\cite{liu2024jailjudge,chen2025radar,kim2024groundial,meade2023using,sahu2025minds}. \textbf{RoTRAG} extends this line of work by using retrieved Rule of Thumb as explicit normative evidence for multi-turn harm detection, together with a lightweight routing mechanism that invokes retrieval-grounded reasoning only when needed.

\section{Methodology}

\subsection{Overview of RoTRAG}

We propose \textbf{RoTRAG}, a retrieval-augmented framework for turn-level Rule of Thumb (RoT) generation under intervention-aware decision making. Given a dialogue up to turn \(i\), the framework first determines whether the current turn requires additional RoT generation. A fine-tuned lightweight routing classifier takes the current turn context and previous RoT as input, and predicts whether to reuse the prior RoT or forward the turn to RoT generation. If the turn is judged not to require additional RoT generation, the system outputs a \textbf{Pass} decision. Otherwise, it retrieves relevant RoT examples from a retrieval corpus and generates a turn-specific RoT conditioned on the retrieved evidence. Overall, the framework consists of three main components: (1) turn-level classification and routing, (2) retrieval-augmented RoT generation, and (3) prediction from accumulated RoT history.

\subsection{Retrieval Corpus Representation}

For retrieval-grounded RoT generation, we use the \texttt{action} and \texttt{RoT} fields from the retrieval corpus. Each corpus item is represented as
\[
d_j = (\mathrm{action}_j, \mathrm{RoT}_j),
\]
where \(\mathrm{action}_j\) denotes the person's behavior or utterance in a social situation and \(\mathrm{RoT}_j\) denotes the corresponding RoT text.

To build the retrieval index, we directly encode the raw action text rather than introducing an additional summarization step. We adopt this design because the action itself captures the core behavioral signal of each instance and serves as the most direct key for identifying relevant RoT examples. For each corpus item, we compute
\[
v_j = \mathrm{Emb}(\mathrm{action}_j),
\]
where \(v_j\) denotes the embedding of \(\mathrm{action}_j\). The collection of these embeddings
\[
\mathcal{V} = \{v_j\}_{j=1}^{N}
\]
is stored as the retrieval index, where \(N\) is the size of the retrieval corpus. This design keeps corpus construction simple and efficient, avoiding unnecessary preprocessing while enabling effective nearest-neighbor retrieval over semantically related action--RoT pairs.

\subsection{Turn-Level Classification and Routing}

\paragraph{Fine-Tuning for Routing Classification}
For routing classification, each training instance is serialized into a single input sequence consisting of the previous RoT, current context, and current response. The binary labels \(z_i \in \{0,1\}\) are obtained by first collecting human annotations on a seed set and then extending the annotations using an LLM reasoning model guided by a high-performing, empirically validated prompt.\footnote{Further details on the prompt validation are provided in Section~4.2.} Using this supervision data, we train the routing classifier as an encoder-based sequence classification model with the standard cross-entropy loss:
\[
\mathcal{L}_{\mathrm{cls}} = - \sum_{c=1}^{2} y_c \log p_c.
\]

\paragraph{Classification and Routing}
For each turn, \textbf{RoTRAG} first determines whether additional RoT generation is necessary. Rather than generating a new RoT for every turn, the framework uses a routing mechanism that selectively invokes the RoT generation module only when the current turn is predicted to require it.

At turn \(i\), the routing classifier takes as input the current turn context, \(C_i\), together with the RoT generated for the previous turn, \(\mathrm{RoT}_{i-1}\), and predicts
\[
\hat{z}_i = f_{\mathrm{cls}}(\mathrm{RoT}_{i-1}, C_i),
\]
where \(\hat{z}_i \in \{0,1\}\). Here, \(\hat{z}_i = 1\) indicates that the current turn can be passed without additional RoT generation, whereas \(\hat{z}_i = 0\) indicates that the turn should be forwarded to the retrieval-augmented RoT generation module. For the first turn, where no previous RoT is available, we directly invoke the RoT generation stage.

This design allows the classifier to assess the current turn not only based on its local dialogue content but also in light of the RoT generated for the immediately preceding turn. As a result, the routing decision reflects both the current utterance and the recent RoT context accumulated during the conversation. From a probabilistic perspective, the classifier estimates
\[
P(Z_i \mid \mathrm{RoT}_{i-1}, C_i),
\]
where \(Z_i\) indicates whether additional RoT generation is required for turn \(i\).

The label mapping in Figure~\ref{fig:ours_framework} is illustrative, while actual predictions follow the dataset-specific label space of each benchmark.\footnote{Detailed routing classifier performance on the validation set and corresponding case studies are provided in Section 7.2 and our \href{https://github.com/GitLeo1/RoTRAG-Rule-of-Thumb-Reasoning-for-Conversation-Harm-Detection-with-Retrieval-Augmented-Generation/blob/main/appendix/routing-classifier-case-study.md}{Github-routing-classifier-case-study.md}.}

\begin{algorithm}[h]

\caption{\emph{RoTRAG inference for multi-turn harm assessment}}
\label{alg:rotrag}
\begin{algorithmic}[1]
\Require Dialogue turns $C_{\leq T}$, indexed retrieval corpus $\mathcal{D}=\{d_j\}_{j=1}^{N}$ with corpus embeddings $\mathcal{V}=\{v_j\}_{j=1}^{N}$, embedding model $\mathrm{Emb}(\cdot)$, routing classifier $f_{\mathrm{cls}}$, turn summarizer $g_{\mathrm{sum}}$, RoT generator $g_{\mathrm{rot}}$, prediction model $h$, top-$k$
\Ensure Predicted labels $\{\hat{y}_i\}_{i=1}^{T}$

\State Initialize RoT history $\mathcal{H}_{\mathrm{RoT}} \leftarrow \emptyset$
\For{$i = 1$ to $T$}
    \If{$i = 1$}
        \State Summarize current turn $s_i \leftarrow g_{\mathrm{sum}}(C_i)$
        \State Encode query $q_i \leftarrow \mathrm{Emb}(s_i)$
        \State Retrieve $\mathcal{N}_i \leftarrow \mathrm{TopK}(q_i, \mathcal{V}, k)$
        \State Set $\mathcal{R}_i \leftarrow \{d_j \mid j \in \mathcal{N}_i\}$
        \State Generate $\mathrm{RoT}_i \leftarrow g_{\mathrm{rot}}(C_i, \mathcal{R}_i)$
    \Else
        \State Predict $\hat{z}_i \leftarrow f_{\mathrm{cls}}(\mathrm{RoT}_{i-1}, C_i)$
        \If{$\hat{z}_i = 1$}
            \State \textbf{Pass} and reuse $\mathrm{RoT}_{i-1}$
            \State $\mathrm{RoT}_i \leftarrow \mathrm{RoT}_{i-1}$
        \Else
            \State Summarize current turn $s_i \leftarrow g_{\mathrm{sum}}(C_i)$
            \State Encode query $q_i \leftarrow \mathrm{Emb}(s_i)$
            \State Retrieve $\mathcal{N}_i \leftarrow \mathrm{TopK}(q_i, \mathcal{V}, k)$
            \State Set $\mathcal{R}_i \leftarrow \{d_j \mid j \in \mathcal{N}_i\}$
            \State Generate $\mathrm{RoT}_i \leftarrow g_{\mathrm{rot}}(C_i, \mathcal{R}_i)$
        \EndIf
    \EndIf
    \State Update $\mathcal{H}_{\mathrm{RoT}} \leftarrow \mathcal{H}_{\mathrm{RoT}} \cup \{\mathrm{RoT}_i\}$
    \State Predict $\hat{y}_i \leftarrow h(\mathcal{H}_{\mathrm{RoT}}, C_{\leq i})$
\EndFor
\State \Return $\{\hat{y}_i\}_{i=1}^{T}$
\end{algorithmic}
\end{algorithm}

\subsection{Retrieval-Augmented RoT Generation}

When a turn is routed to the RoT generation stage (\(\hat{z}_i = 0\)), the framework first summarizes the current turn using an LLM:
\[
s_i = g_{\mathrm{sum}}(C_i),
\]
where \(s_i\) denotes the summary of the current turn context \(C_i\). This summarization step is intended to provide a more compact query representation by reducing irrelevant conversational detail and bringing the current turn closer to the action-oriented representation used in the retrieval corpus.

The summary is then encoded into a query embedding:
\[
q_i = \mathrm{Emb}(s_i).
\]
Based on \(q_i\), the system identifies the indices of the top-\(k\) nearest corpus items:
\[
\mathcal{N}_i = \mathrm{TopK}(q_i, \mathcal{V}, k),
\]
where \(\mathcal{V} = \{v_j\}_{j=1}^{N}\) denotes the indexed corpus embeddings. The retrieved set is then defined as
\[
\mathcal{R}_i = \{ d_j \mid j \in \mathcal{N}_i \}.
\]

The retrieved items correspond to action--RoT examples that are semantically similar to the current turn and thus provide relevant external evidence for RoT generation. Conditioned on the current turn context and the retrieved examples, the model generates the RoT for turn \(i\):
\[
\mathrm{RoT}_i = g_{\mathrm{rot}}(C_i, \mathcal{R}_i).
\]

In this way, the retrieval corpus functions as a structured memory of prior action--RoT patterns, helping the model generate more consistent and contextually appropriate RoTs than zero-shot generation alone.

\subsection{Prediction from Accumulated RoT History}

As shown in Figure~\ref{fig:ours_framework}, the generated RoTs are accumulated and used to support final label prediction. Specifically, the RoT history up to turn \(i\), together with the dialogue context up to turn \(i\), is provided to a prediction module that outputs the final label:
\[
\hat{y}_i = h(\mathrm{RoT}_{\leq i}, C_{\leq i}),
\]
where \(h\) denotes the label prediction module.

In this setting, \(\hat{y}_i\) belongs to the dataset-specific label space defined by the corresponding benchmark. Predictions are evaluated against the ground truth using standard classification metrics. Under this formulation, RoTs are not merely explanatory text; they serve as intermediate reasoning representations that support the final harm or severity prediction task. Algorithm~\ref{alg:rotrag} summarizes the overall inference procedure of \textbf{RoTRAG}, integrating the routing, retrieval, RoT generation, and final prediction steps described above.

\section{Experiment Setup}

\subsection{Models Used}
For the main experiments, we use both API-based and open-source models. Specifically, \textit{GPT-4o mini}~\cite{openai2024gpt} is chosen as a representative API-based model, and \textit{Qwen3-14B}~\cite{yang2025qwen3} is selected as the open-source LLM backbone, with temperature $= 0$ for deterministic outputs. For retrieval, we adopt a widely used \textit{e5-base}~\cite{wang2022text} model as the embedding model to encode action representations and perform nearest-neighbor search over the retrieval corpus, retrieving the top-5 most relevant RoTs as context. For turn-level classification, we use \textit{RoBERTa-large}~\cite{liu2019roberta} as the encoder-based classifier.\footnote{Detailed comparisons among candidate classifiers, along with the hyperparameter settings of the final \textit{RoBERTa-large} classifier, are provided in Section 6.3.} We additionally include \textit{GPT-5.4-Thinking}~ \footnote{\url{https://deploymentsafety.openai.com/gpt-5-4-thinking}}, \textit{Claude 3.7 Sonnet (thinking)}~ \cite{anthropic2025claude37}, and \textit{DeepSeek-V3.2 (thinking)}~\cite{liu2025deepseek} as extended reasoning baselines. To generate the training data used for routing classifier supervision, we employ \textit{GPT-o4 mini}, a reasoning-capable model, to produce synthetic annotations under the prompt design.

\subsection{Datasets}
\label{sec:datasets}
\begin{table}[t]
\centering
\small
\setlength{\tabcolsep}{3.5pt}
\renewcommand{\arraystretch}{1.12}
\caption{Summary of all datasets used in the experiments. D-Rules-of-Thumb is used as the retrieval corpus. ProsocialDialog train/valid is used for RoT generation and classifier training, while ProsocialDialog test and Safety Reasoning Multi-Turn Dialogue are used for evaluation.}
\label{tab:main_dataset_summary}
\begin{tabular}{lccc}
\toprule
\textbf{Dataset} & \textbf{Role} & \textbf{Target} & \textbf{Size} \\
\midrule
D-Rules-of-Thumb & Retrieval & -- & 577.9k \\
ProsocialDialog  & Train/Valid & RoT / Cls. & 120k / 20.4k \\
ProsocialDialog  & Test & \texttt{safety\_label} & 25k \\
Safety Reasoning MTD & Test & \texttt{q\_sev}, \texttt{r\_sev} & 6.46k \\
\bottomrule
\end{tabular}
\end{table}

Table~\ref{tab:main_dataset_summary} summarizes all datasets used in this work, including the retrieval corpus, training/validation data, and test benchmarks.

\paragraph{Retriever Corpus} For retrieval, we use the \textit{D-Rules-of-Thumb}~\cite{rao2023makes} dataset, which contains structured action--RoT pairs. In our framework, the \texttt{action} field is used as the retrieval key, and the paired \texttt{RoT} field is used as the corresponding reasoning target.

\paragraph{Training Dataset} For training, we use the training split of the \textit{Prosocial}~\cite{kim2022prosocialdialog} dataset. To construct high-quality supervision labels, we first recruited ten human annotators and collected RoT labels for 1{,}000 training instances. The final ground-truth label for each instance was determined by hard voting across annotators. Based on these human-annotated examples, we designed a prompt and validated it against the human ground truth. The prompt achieved 98\% accuracy with respect to the human-labeled annotations using GPT-o4 mini, and we then used this prompt to generate labels for the remaining training instances.

\paragraph{Test Dataset}
For evaluation, we use two benchmark test sets. On \textit{ProsocialDialog}, the model predicts \texttt{safety\_label} from \texttt{context} and \texttt{response}, where labels are defined on a 1--5 severity scale. 
On \textit{Safety Reasoning Multi-Turn Dialogue}~\cite{kuo2025safety}, the model predicts two labels, \texttt{question\_severity} and \texttt{response\_severity}, from the full \texttt{conversation}, where severity scores range from 0 to 10.
Together, these datasets evaluate the framework under both prosocial safety classification and multi-turn severity reasoning settings.\footnote{Detailed labeling process and dataset descriptions are provided in our \href{https://github.com/GitLeo1/RoTRAG-Rule-of-Thumb-Reasoning-for-Conversation-Harm-Detection-with-Retrieval-Augmented-Generation/blob/main/appendix/annotation-and-datasets.md}{Github-annotation-and-datasets.md}.}

\begin{table*}[t]
\centering
\caption{Main results on the \textit{Prosocial} and \textit{Safety} benchmarks under GPT-4o mini and QWEN3-14B. Accuracy, Precision, Recall, and F1 are reported for RoTRAG and all baselines. For Safety, results are shown separately for Safety Question and Safety Response, together with their aggregate Safety Overall. \textbf{Bold} indicates the best result and \underline{underline} the second-best result within each backbone group.}
\resizebox{\linewidth}{!}{%
\begin{tabular}{l cccc cccc cccc cccc}
\toprule
 & \multicolumn{4}{c}{Prosocial}
 & \multicolumn{4}{c}{Safety Question}
 & \multicolumn{4}{c}{Safety Response}
 & \multicolumn{4}{c}{Safety Overall} \\
\cmidrule(lr){2-5}\cmidrule(lr){6-9}\cmidrule(lr){10-13}\cmidrule(lr){14-17}
 & Acc & Prec & Rec & F1
 & Acc & Prec & Rec & F1
 & Acc & Prec & Rec & F1
 & Acc & Prec & Rec & F1 \\
\midrule
\rowcolor{gray!10}\multicolumn{17}{l}{\textbf{GPT-4o mini}}\\

Zero Shot
  & 0.3720             & \underline{0.4541} & \underline{0.3720} & 0.3217
  & 0.2373 & 0.1317          & 0.1369            & 0.0981
  & 0.2649             & 0.1346          & 0.1273            & 0.0969
  & 0.2611             & 0.1332          & 0.1321            & 0.0975 \\
Chain-of-Thought (CoT)~(\citeyear{wei2022chain}) 
  & 0.3634             & 0.4471          & 0.3634            & \underline{0.3565}
  & 0.2455             & \underline{0.1766} & 0.1565          & 0.1068
  & 0.2367             & 0.1174          & 0.1172            & 0.0889
  & 0.2411             & 0.1470          & 0.1369            & 0.0979 \\
Self-Consistency~(\citeyear{wangself})
  & 0.2943             & 0.3306          & 0.3487            & 0.2631
  & \underline{0.2487}    & 0.1621          & 0.1470            & 0.1177
  & \textbf{0.2712}    & 0.1351          & 0.1353            & \underline{0.1110}
  & \textbf{0.2650}    & 0.1486          & 0.1412            & 0.1144 \\
Role Assignment~(\citeyear{shanahan2023role}) 
  & 0.2967             & 0.4481          & 0.2967            & 0.2696
  & 0.2364             & 0.1424          & \underline{0.1613} & 0.1114
  & 0.2464             & 0.1370          & 0.1195            & 0.0916
  & 0.2414             & 0.1397          & 0.1404            & 0.1015 \\
JailJudge~(\citeyear{liu2024jailjudge}) 
  & 0.2508             & 0.2910          & 0.2720            & 0.2236
  & 0.2340             & 0.1213          & 0.1580            & \underline{0.1250}
  & 0.2217             & 0.1255          & 0.1378            & 0.1106
  & 0.2279             & 0.1234          & \underline{0.1479} & \underline{0.1178} \\
RADAR~(\citeyear{chen2025radar}) 
  & \textbf{0.4221}    & 0.4239          & 0.3047            & 0.3082
  & 0.2011             & 0.1702          & 0.1352            & 0.1040
  & 0.2000             & \underline{0.1606} & \underline{0.1392} & \underline{0.1110}
  & 0.2006             & \underline{0.1654} & 0.1372          & 0.1075 \\
\textbf{RoTRAG (Ours)}
  & \underline{0.4018} & \textbf{0.4593} & \textbf{0.4018} & \textbf{0.3909}
  & \textbf{0.2570}             & \textbf{0.3301} & \textbf{0.2570}   & \textbf{0.2068}
  & \underline{0.2686} & \textbf{0.2543} & \textbf{0.2686}   & \textbf{0.2019}
  & \underline{0.2628} & \textbf{0.2922} & \textbf{0.2628}   & \textbf{0.2044} \\
\midrule
\rowcolor{gray!10}\multicolumn{17}{l}{\textbf{QWEN3-14B}}\\
Zero Shot
  & 0.3142             & 0.3760          & 0.3142            & 0.2533
  & 0.2112             & 0.1839          & 0.2112            & 0.1645
  & 0.1894             & 0.1076          & 0.1894            & 0.0955
  & 0.2003             & 0.1458          & 0.2003           & \underline{0.1300} \\
Chain-of-Thought (CoT)~(\citeyear{wei2022chain}) 
  & 0.3166             & \underline{0.3784} & 0.3166          & 0.2691
  & 0.2190             & \underline{0.2529} & 0.2190          & 0.1717
  & 0.1809             & 0.0940          & 0.1809            & 0.0828
  & 0.1341             & 0.1585          & 0.1341            & 0.1273 \\
Self-Consistency~(\citeyear{wangself})
  & \underline{0.3471} & 0.3392          & 0.3492            & 0.2882
  & 0.2052             & 0.2471          & 0.2052            & 0.1391
  & \underline{0.2169} & 0.0965          & \underline{0.2169} & \underline{0.1100}
  & \underline{0.2111} & \underline{0.1718} & \underline{0.2111} & 0.1246 \\
Role Assignment~(\citeyear{shanahan2023role}) 
  & 0.3242             & 0.3780          & 0.3242            & \underline{0.3162}
  & \underline{0.2210} & 0.2100          & \underline{0.2210} & \underline{0.1773}
  & 0.2098             & \underline{0.1077} & 0.1998          & 0.0863
  & 0.1483             & 0.1489          & 0.1483            & 0.1218 \\
JailJudge~(\citeyear{liu2024jailjudge}) 
  & 0.2567             & 0.2802          & 0.2676            & 0.2181
  & 0.1941             & 0.1006          & 0.1172            & 0.0752
  & 0.1991             & 0.0893          & 0.1073            & 0.0650
  & 0.2081             & 0.0950          & 0.1123            & 0.0701 \\
RADAR~(\citeyear{chen2025radar}) 
  & 0.3020             & 0.3466          & \underline{0.3538} & 0.3030
  & 0.1770             & 0.1418          & 0.1440            & 0.1046
  & 0.1840             & 0.0990          & 0.1000            & 0.0852
  & 0.1805             & 0.1204          & 0.1220            & 0.0949 \\

\textbf{RoTRAG (Ours)}
  & \textbf{0.3929}    & \textbf{0.4283} & \textbf{0.3929}   & \textbf{0.3635}
  & \textbf{0.2467}    & \textbf{0.2765} & \textbf{0.2467}   & \textbf{0.1908}
  & \textbf{0.2220}    & \textbf{0.1899} & \textbf{0.2320}   & \textbf{0.1723}
  & \textbf{0.2344}    & \textbf{0.1932} & \textbf{0.2344}   & \textbf{0.1416} \\
\midrule
\rowcolor{gray!10}\multicolumn{17}{l}{\textbf{Thinking Models}}\\
GPT-5.4-Thinking~ (2025)
  & 0.3349 & 0.3635 & 0.3349 & 0.2993
  & 0.2519 & 0.3256 & 0.2519 & 0.1615
  & 0.2268 & 0.1796 & 0.2268 & 0.1264
  & 0.2393 & 0.2626 & 0.2393 & 0.1940 \\
Claude 3.7 Sonnet (thinking)~ (2025)
  & 0.2521 & 0.3018 & 0.3003 & 0.2311
  & 0.2029 & 0.2370 & 0.1190 & 0.1171
  & 0.1977 & 0.0910 & 0.0977 & 0.0760
  & 0.2003 & 0.1640 & 0.1084 & 0.0966 \\
DeepSeek-V3.2 (thinking)~ (\citeyear{liu2025deepseek})
  & 0.3391 & 0.2493 & 0.3136 & 0.2660
  & 0.2457 & 0.1575 & 0.1136 & 0.0857
  & 0.2265 & 0.0908 & 0.0991 & 0.0598
  & 0.2361 & 0.1242 & 0.1064 & 0.0728 \\
\bottomrule
\end{tabular}
}
\label{tab:main_results}
\end{table*}

\begin{table}[t]
\centering
\caption{Evaluation results using MAE and distribution-level metrics on the \textit{Prosocial} and \textit{Safety} benchmarks. MAE measures how closely the predicted scores match the target values, while TVD and EMD evaluate how well each method recovers the overall label distribution. Lower is better for all metrics.}
\label{tab:distribution_metrics}
\resizebox{\columnwidth}{!}{%
\begin{tabular}{l|ccc|ccc}
\toprule
\multirow{2}{*}{\textbf{GPT-4o mini}}
& \multicolumn{3}{c|}{\textbf{Prosocial}} 
& \multicolumn{3}{c}{\textbf{Safety (Avg.)}} \\
 & MAE & TVD & EMD & MAE & TVD & EMD \\
\midrule
Zero Shot   
& 0.9392 & 0.4310 & \underline{0.4310} 
& 2.9977 & 0.4989 & 2.6011 \\

Chain-of-Thought (CoT)~(\citeyear{wei2022chain})        
& 0.8948 & 0.4062 & 0.4737 
& 3.0455 & 0.5375 & 2.6486 \\

Self Consistency~(\citeyear{wangself})
& 1.2748 & 0.4392 & 0.9452 
& \underline{2.8393} & \underline{0.4639} & \underline{2.5412} \\

Role Assignment~(\citeyear{shanahan2023role}) 
& 0.9319 & 0.5293 & 0.6254 
& 3.0630 & 0.5290 & 2.7582 \\

JailJudge~(\citeyear{liu2024jailjudge})  
& 1.4022 & 0.4166 & 0.9338 
& \textbf{2.6912} & 0.4992 & 2.9044 \\

RADAR~(\citeyear{chen2025radar})     
& \underline{0.8914} & \underline{0.3012} & 0.4542 
& 3.1949 & 0.6350 & 2.7442 \\

RoTRAG (Ours)
& \textbf{0.8699} & \textbf{0.2468} & \textbf{0.4078} 
& 2.8949 & \textbf{0.4482} & \textbf{2.3513} \\

\midrule
\multirow{2}{*}{\textbf{QWEN3-14B}}
& \multicolumn{3}{c|}{\textbf{Prosocial}} 
& \multicolumn{3}{c}{\textbf{Safety (Avg.)}} \\
 & MAE & TVD & EMD & MAE & TVD & EMD \\
\midrule

Zero Shot  
& 1.0890 & 0.4100 & 0.9217 
& 3.3259 & 0.6143 & 2.8898 \\

Chain-of-Thought (CoT)~(\citeyear{wei2022chain})         
& \underline{0.9238} & 0.4146 & \underline{0.4855} 
& 3.3024 & \underline{0.5126} & 2.8436 \\

Self Consistency~(\citeyear{wangself}) 
& 0.9809 & 0.4264 & 0.5000 
& 3.0262 & 0.6089 & 2.8299 \\

Role Assignment~(\citeyear{shanahan2023role})
& 0.9286 & 0.4140 & 0.5275 
& 3.4211 & 0.5633 & \underline{2.6809} \\

JailJudge~(\citeyear{liu2024jailjudge}) 
& 1.6217 & 0.4392 & 1.2417 
& 3.1074 & 0.5603 & 2.7105 \\

RADAR~(\citeyear{chen2025radar})      
& 1.1140 & \underline{0.3656} & 0.7260 
& \underline{2.9339} & 0.5445 & 3.1921 \\

RoTRAG (Ours)
& \textbf{0.8855} & \textbf{0.3455} & \textbf{0.4510} 
& \textbf{2.8189} & \textbf{0.4745} & \textbf{2.3427} \\

\bottomrule
\end{tabular}%
}
\end{table}

\subsection{Baselines}
We compare our method against a diverse set of prompting, safety-oriented, multi-agent, and thinking-model baselines, covering standard single-pass prompting, reasoning-based prompting, and collaborative decision-making settings. The thinking models are evaluated in a zero-shot setting.

\begin{itemize}
    \item \textbf{Single-Inference Baselines}
    \begin{itemize}
        \item \textit{Zero-shot}: A direct prompting baseline that predicts the safety label from the dialogue context without explicit intermediate reasoning.
        
        \item \textit{Chain-of-Thought (CoT)}~\citep{wei2022chain}: A reasoning-based prompting baseline that encourages the model to generate step-by-step justifications before making a final prediction.
        
        \item \textit{Role Assignment}~\citep{shanahan2023role}: A prompting baseline that improves judgment consistency by assigning the model a specific evaluative role during safety assessment.
    \end{itemize}
    
    \item \textbf{Multi-Agent / Multi-Inference Baselines}
    \begin{itemize}
        \item \textit{Self-consistency}~\citep{wangself}: A reasoning baseline that samples multiple reasoning paths and aggregates them to produce a more stable final decision.
        
        \item \textit{JailJudge}~\citep{liu2024jailjudge}: A safety-focused judging framework designed to detect harmful or unsafe content through structured model-based evaluation.
        
        \item \textit{RADAR}~\citep{chen2025radar}: A recent safety baseline that performs harm assessment using a more advanced reasoning and risk-detection framework.
    \end{itemize}
    
    \item \textbf{Thinking Models}
    \begin{itemize}
        \item \textit{GPT-5.4 Thinking}\footnote{\url{https://deploymentsafety.openai.com/gpt-5-4-thinking}}: An OpenAI reasoning model used to evaluate the zero-shot safety judgment capability of a recent proprietary thinking model.
        
        \item \textit{Claude 3.7 Sonnet}~\cite{anthropic2025claude37}: A hybrid reasoning model with extended thinking, included to compare safety judgment performance across a different proprietary model family.
        
        \item \textit{DeepSeek-V3.2}~\cite{liu2025deepseek}: An open-weight reasoning-oriented model used to examine whether strong reasoning and agent-oriented capabilities can support robust harm assessment without task-specific training.
    \end{itemize}
\end{itemize}

These models are included as reasoning-focused reference models to assess how far intrinsic reasoning ability alone can support zero-shot conversation harm detection.\footnote{Detailed baseline definitions and implementation settings are provided in our \href{https://github.com/GitLeo1/RoTRAG-Rule-of-Thumb-Reasoning-for-Conversation-Harm-Detection-with-Retrieval-Augmented-Generation/blob/main/appendix/baselines-and-hardware.md}{Github-baselines-and-hardware.md}.}

\begin{table*}[h]
\centering
\caption{Ablation results on the Prosocial and Safety benchmarks for GPT-4o mini and QWEN3-14B: (1) Full RoT Generation, and (2) Random Routing. \textbf{Bold} denotes the best result and \underline{underline} the second best within each backbone group.}
\resizebox{\linewidth}{!}{%
\begin{tabular}{ll cccc cccc cccc cccc}
\toprule
 & & \multicolumn{4}{c}{Prosocial}
   & \multicolumn{4}{c}{Safety Question}
   & \multicolumn{4}{c}{Safety Response}
   & \multicolumn{4}{c}{Safety Overall} \\
\cmidrule(lr){3-6}\cmidrule(lr){7-10}\cmidrule(lr){11-14}\cmidrule(lr){15-18}
 & & Acc & Prec & Rec & F1
   & Acc & Prec & Rec & F1
   & Acc & Prec & Rec & F1
   & Acc & Prec & Rec & F1 \\
\midrule
\rowcolor{gray!10}\multicolumn{18}{l}{\textbf{GPT-4o mini}}\\
& Full RoT Generation
& 0.3752 & 0.3648 & 0.3612 & 0.3021
& \underline{0.2345} & \underline{0.2643} & \underline{0.1489} & \underline{0.1264}
& \underline{0.2515} & \underline{0.1258} & \underline{0.1383} & \underline{0.1035}
& \underline{0.2430} & \underline{0.1950} & \underline{0.1436} & \underline{0.1149} \\

& Random Routing
& \underline{0.4003} & \underline{0.3899} & \underline{0.3835} & \underline{0.3350}
& 0.2321 & 0.2429 & 0.1439 & 0.1207
& 0.2481 & 0.1193 & 0.1344 & 0.0999
& 0.2401 & 0.1811 & 0.1392 & 0.1103 \\

& \textbf{RoTRAG (Ours)}
& \textbf{0.4018} & \textbf{0.4593} & \textbf{0.4018} & \textbf{0.3909}
& \textbf{0.2570} & \textbf{0.3301} & \textbf{0.2570} & \textbf{0.2068}
& \textbf{0.2686} & \textbf{0.2543} & \textbf{0.2686} & \textbf{0.2019}
& \textbf{0.2628} & \textbf{0.2922} & \textbf{0.2628} & \textbf{0.2044} \\
\midrule
\rowcolor{gray!10}\multicolumn{18}{l}{\textbf{QWEN3-14B}}\\
& Full RoT Generation
& 0.3752 & 0.3404 & 0.3280 & 0.2605
& \underline{0.2402} & 0.2273 & \underline{0.1269} & 0.1022
& \underline{0.2144} & \underline{0.0730} & \underline{0.0952} & \underline{0.0426}
& \underline{0.2273} & 0.1502 & \underline{0.1111} & 0.0724 \\

& Random Routing
& \textbf{0.3936} & \underline{0.3477} & \underline{0.3490} & \underline{0.2852}
& 0.2322 & \underline{0.2384} & \underline{0.1269} & \underline{0.1058}
& 0.2131 & 0.0625 & 0.0944 & 0.0413
& 0.2227 & \underline{0.1505} & 0.1107 & \underline{0.0736} \\

& \textbf{RoTRAG (Ours)}
& \underline{0.3929} & \textbf{0.4283} & \textbf{0.3929} & \textbf{0.3635}
& \textbf{0.2467} & \textbf{0.2765} & \textbf{0.2467} & \textbf{0.1908}
& \textbf{0.2220} & \textbf{0.1899} & \textbf{0.2320} & \textbf{0.1723}
& \textbf{0.2344} & \textbf{0.1932} & \textbf{0.2344} & \textbf{0.1416} \\
\bottomrule
\end{tabular}
}
\label{tab:ablation}
\end{table*}

\subsection{Evaluation Metrics}

We evaluate all methods using classification, value-based, and distribution based metrics. For classification performance, we report \textit{Accuracy}, \textit{Precision}, \textit{Recall}, and \textit{F1} over the dataset-specific target labels. Accuracy measures overall correctness, while Precision, Recall, and F1 provide a more detailed view of classification quality, especially under label imbalance. For multi-class settings, we report the corresponding scores over the dataset-specific label space.

To complement these classification metrics, we additionally evaluate whether predictions remain close to the target values and preserve the overall label distribution. For this purpose, we report Mean Absolute Error (MAE), Total Variation Distance (TVD), and Earth Mover's Distance (EMD). MAE measures the average difference between predicted and target scores at the instance level, while TVD and EMD assess how closely the predicted label distribution matches the ground-truth distribution. Lower values indicate better performance for all three metrics.

Together, these metrics capture classification quality, score closeness, and distributional alignment.

\section{Experimental Results}
\subsection{Main Experiments}

As shown in Table~\ref{tab:main_results}, we compare RoTRAG with a diverse set of baselines, including direct prompting, reasoning-based prompting, role assignment, multi-inference methods, and safety-focused judging frameworks. While some baselines achieve competitive results in specific settings, their performance is not consistent across datasets and backbones, largely because they rely mainly on the model’s internal reasoning ability. Notably, even state-of-the-art thinking models underperform on our tasks, suggesting that advanced intrinsic reasoning alone is insufficient for reliable harm detection. The ambiguity and social nuance of conversation-level safety judgments require alignment with external, shared norms, which pure reasoning does not guarantee.

In contrast, our framework \textbf{RoTRAG} adopts a retrieval-augmented generation framework that strengthens risk assessment by incorporating external knowledge, while also introducing RoT to provide socially and ethically appropriate guidance. Rather than making a surface-level prediction of whether a dialogue is risky, \textbf{RoTRAG} interprets risk intensity by jointly considering external evidence and normative cues relevant to the dialogue context. In other words, \textbf{RoTRAG} does not rely solely on the language model’s pretrained knowledge and internal reasoning ability, but instead performs more grounded judgment by leveraging both retrieved information and normative guidance. This design leads to predictions that are not only more reasonable but also more accurate than those of purely reasoning-based approaches, and we believe it is a key factor behind the consistent performance gains observed across multiple datasets and backbone models.

This advantage appears in both the main classification results and the additional score- and distribution-level evaluations. Compared with the strongest baseline in each setting, \textbf{RoTRAG} improves F1 by 9.6\%--81.9\% with GPT-4o mini and by 7.6\%--56.6\% with QWEN3-14B, with especially large gains on Safety-related settings. Beyond accuracy-based evaluation, we further report MAE, TVD, and EMD (Table~\ref{tab:distribution_metrics}), which assess prediction closeness and distributional alignment. Under these metrics, \textbf{RoTRAG} achieves the best result in 11 of 12 settings, with an average relative gain of about 7\% over the strongest competing baseline where it ranks first. Overall, these results show that \textbf{RoTRAG} not only improves classification performance, but also produces predictions that are closer to the target values and better aligned with the ground-truth distribution.

\subsection{Ablation Study}

To analyze the contribution of the routing component in \textbf{RoTRAG}, we evaluate two ablation variants: (1) \textit{Full RoT Generation}, which generates RoTs for all turns; and (2) \textit{Random Routing}, which applies random routing with the same generation ratio as the classifier. The results are shown in Table~\ref{tab:ablation}.

\subsubsection{Full RoT Generation}
Full RoT Generation removes the routing classifier and generates a new RoT for every turn. This tests whether simply increasing the amount of intermediate reasoning improves performance. However, generating RoTs for all turns introduces redundant or weakly relevant norms, which can dilute the main risk signal and reduce prediction quality. This pattern is consistent across both backbones, showing that selective filtering is important not only for efficiency but also for maintaining a relevant RoT history.

\subsubsection{Random Routing}
Random Routing tests whether the gain of the full model comes from the learned routing decisions or simply from reducing the number of generated RoTs. To control for this, we preserve the same 0/1 ratio\footnote{Detailed classifier prediction ratios on the test set are provided in the Prediction Distribution section of our \href{https://github.com/GitLeo1/RoTRAG-Rule-of-Thumb-Reasoning-for-Conversation-Harm-Detection-with-Retrieval-Augmented-Generation}{Github-README.md}.} as the trained classifier, but randomly assign the routing labels across turns. We report the average over three fixed random seeds (13, 42, and 123). Despite matching the amount of RoT generation, this variant generally performs worse than the full model, showing that the benefit of routing lies in identifying which turns truly require new normative reasoning.

Taken together, the ablation results show that selective routing is crucial to \textbf{RoTRAG}. Generating RoTs for every turn can introduce redundant or weakly relevant signals, while random routing with the same generation ratio underperforms the full model in most metrics. This indicates that the improvement comes not merely from reducing RoT generation, but from learning when new RoT reasoning is needed, helping maintain a more relevant RoT history across benchmarks and backbones.

\section{Discussion}
\subsection{Qualitative Analysis}
To examine the quality of the reasoning produced by \textbf{RoTRAG}, we conducted a qualitative evaluation with five experts in LLM and AI systems. 
The experts assessed 10 examples from two perspectives: 
(H1) how reasonable the newly generated Rule of Thumb (RoT) is on a 1--5 scale, and 
(H2) whether CoT reasoning or \textbf{RoTRAG}'s reasoning better supports the final output.

\begin{figure}[h]
  \centering
  \includegraphics[width=\linewidth]{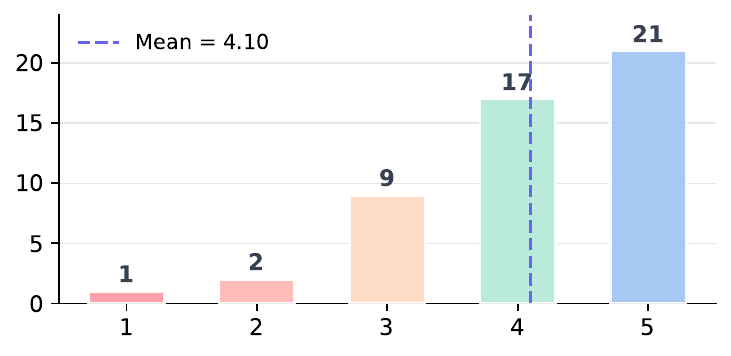}
  \caption{Distribution of expert ratings for the reasonableness of the generated RoTs. Ratings are concentrated at 4 and 5, with a mean of 4.10.}
  \label{fig:qual_score_distribution}
\end{figure}

\begin{table}[h]
\centering
\small
\setlength{\tabcolsep}{10pt}
\renewcommand{\arraystretch}{1.15}
\caption{Preference votes in the qualitative evaluation.}
\label{tab:qual_pref}
\begin{tabular}{lcc}
\toprule
\textbf{Method} & \textbf{Votes} & \textbf{Ratio} \\
\midrule
CoT reasoning & 13 & 26.0\% \\
\textbf{RoTRAG reasoning} & \textbf{37} & \textbf{74.0\%} \\
\bottomrule
\end{tabular}
\end{table}

As shown in Figure~\ref{fig:qual_score_distribution}, \textbf{RoTRAG} produces generally well-received reasoning. 
Across 50 expert ratings, the generated RoTs achieved an average score of 4.10/5, with 76.0\% of all ratings falling in the 4--5 range. 
This indicates that the generated RoTs are generally viewed as reasonable. In addition, the preference summary in Table~\ref{tab:qual_pref} shows that experts selected \textbf{RoTRAG}'s reasoning in 74.0\% of the judgments, while CoT reasoning was preferred in 26.0\%. 
These results indicate that the intermediate reasoning introduced by \textbf{RoTRAG} is not only plausible to human experts, but also more effective in supporting the final output than standard CoT reasoning.

Overall, these findings suggest that \textbf{RoTRAG} improves both predictive performance and perceived reasoning quality.

\subsection{Cost \& Latency Analysis}

\paragraph{Token-Performance Trade-off Analysis}
\begin{figure}[h]
  \centering
  \includegraphics[width=\linewidth]{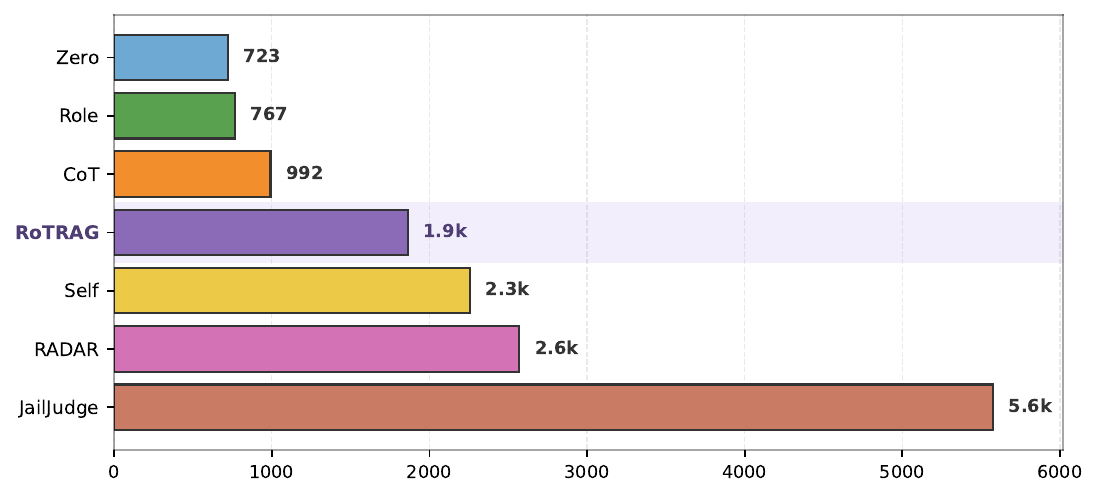}
  \caption{Average token usage of each method on the Prosocial and Safety benchmarks under GPT-4o mini and QWEN3-14B.}
  \label{fig:token_usage_barh}
\end{figure}

\begin{figure}[h]
  \centering
  \includegraphics[width=\linewidth]{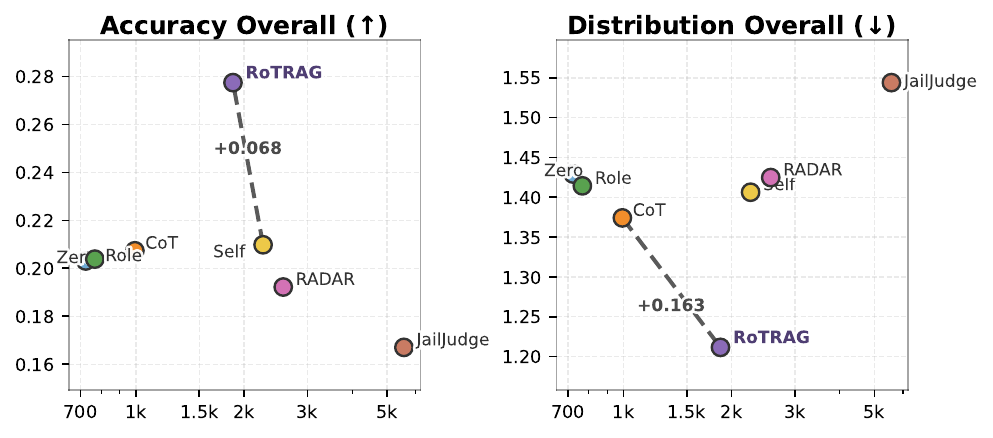}
  \caption{Token-performance trade-off across methods on the Prosocial and Safety benchmarks under GPT-4o mini and QWEN3-14B.}
  \label{fig:token_performance}
\end{figure}

To examine efficiency, we analyze the trade-off between token cost and predictive performance across all baselines and \textbf{RoTRAG}, using overall averages from the Prosocial and Safety datasets under both GPT-4o mini and QWEN3-14B. For the accuracy-based view, \textit{Accuracy Overall} is defined as the average of Accuracy, Precision, Recall, and F1. For the distribution-based view, \textit{Distribution Overall} is defined as the average of TVD and EMD, where lower values indicate better alignment.

Figure~\ref{fig:token_usage_barh} compares the average token usage of each method. Lightweight prompting methods such as Zero-shot, Role Assignment, and CoT require relatively few tokens, whereas more complex methods such as JailJudge and RADAR are substantially more expensive. \textbf{RoTRAG} lies between these extremes, using more tokens than simple prompting baselines while remaining much more efficient than the most expensive multi-step methods.

Figure~\ref{fig:token_performance} shows that \textbf{RoTRAG} achieves the best overall classification performance and the lowest overall distribution error while maintaining moderate token cost. This suggests that its gains come from a more favorable token-performance balance rather than from higher token usage alone.

\paragraph{Time Efficiency}
\begin{figure}[t]
  \centering
  \includegraphics[width=\linewidth]{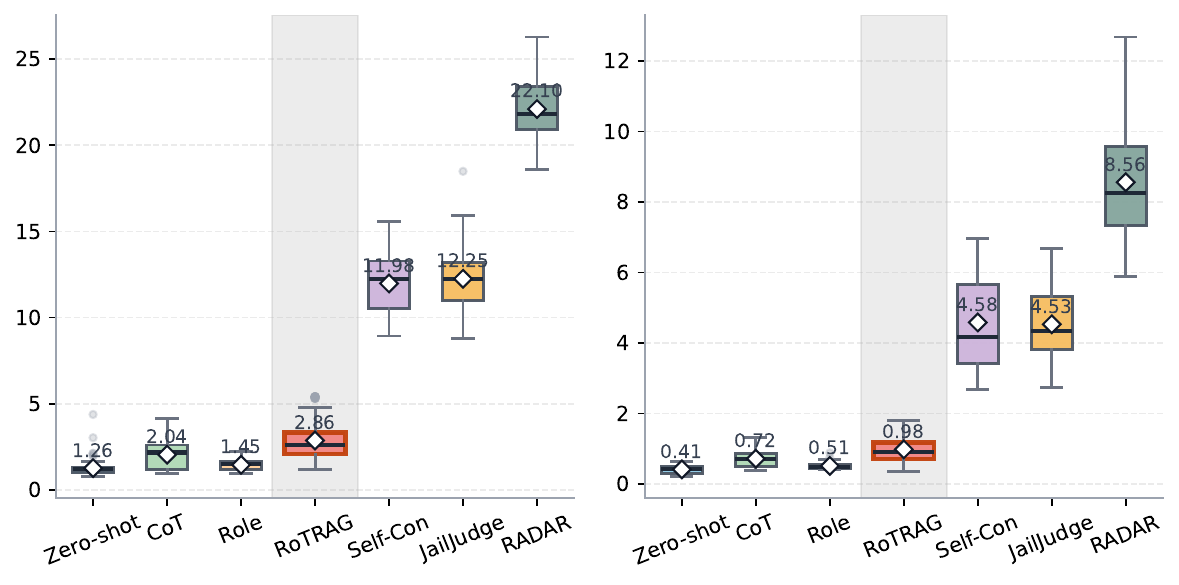}
  \caption{Average time required to generate one prediction for each method. The left panel shows results under GPT-4o mini, and the right panel shows results under QWEN.}
  \label{fig:time_performance}
\end{figure}

Figure~\ref{fig:time_performance} summarizes the inference time of each method under GPT and QWEN. Overall, \textbf{RoTRAG} shows a middle-ground efficiency profile: it is slower than simple single-prompt baselines because it includes routing and retrieval, but it remains substantially faster than multi-agent methods that require repeated generation or interaction.

This pattern indicates that \textbf{RoTRAG} offers a practical balance between effectiveness and efficiency. Although it is not as lightweight as the simplest prompting approaches, it avoids the heavy latency of high-cost multi-agent baselines while delivering stronger overall performance.

\subsection{Routing Classifier Analysis}

The routing classifier is a lightweight binary module that decides whether \textbf{RoTRAG} should generate a new RoT for the current turn (\textit{regenerate}) or reuse the existing RoT history (\textit{reuse}). This design targets a practical tension in multi-turn safety assessment: regenerating RoTs at every turn increases cost and may introduce redundant or weakly relevant norms, while over-reusing can anchor subsequent predictions to an outdated normative frame when the dialogue intent, target, or severity shifts.

\begin{figure}[t]
  \centering
  \includegraphics[width=0.6\linewidth]{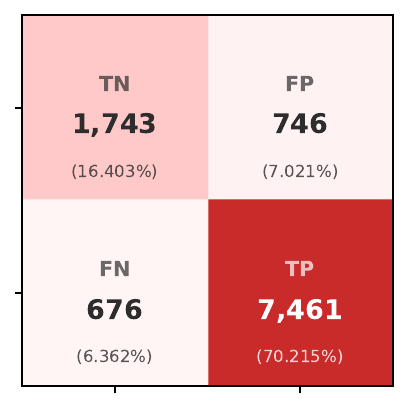}
  \caption{Validation-set confusion matrix of the routing classifier.}
  \label{fig:confusion_matrix}
\end{figure}

\begin{itemize}
  \item \textbf{Training data construction.} We construct turn-level supervision by pairing each turn with its preceding dialogue context and the most recently generated RoT. A turn is labeled as \textit{regenerate} when the new utterance introduces a substantive shift in the normative focus (e.g., a change in the harmful action, target, intent, or severity) such that the previous RoT is no longer sufficient; otherwise it is labeled as \textit{reuse}. This formulation trains the router to detect when additional retrieval-grounded evidence is likely to improve downstream judgment, rather than simply maximizing the number of generated RoTs.
  \item \textbf{Training setup and results.} We fine-tune RoBERTa-large for sequence classification for 3 epochs using AdamW with a linear learning-rate schedule (no warm-up), learning rate $2 \times 10^{-5}$, weight decay 0.01, and max sequence length 256. We use a per-device batch size of 8 with gradient accumulation over 2 steps (effective batch size 16) and enable FP16. Using seed 42 and selecting the best checkpoint by validation macro F1, the classifier reaches 0.8662 validation accuracy at epoch~3. Figure~\ref{fig:confusion_matrix} summarizes the corresponding confusion matrix.
\end{itemize}

\begin{table*}[t]
\centering
\small
\renewcommand{\arraystretch}{1.15}
\caption{Representative validation examples illustrating correct and incorrect routing decisions of the classifier.}
\label{tab:classifier_examples}
\begin{tabular}{c p{0.25\linewidth} p{0.45\linewidth} c c}
\toprule
\textbf{Type} & \textbf{Previous RoT} & \textbf{Current dialogue} & \textbf{GT} & \textbf{Pred.} \\
\midrule

TP &
\begin{minipage}[t]{\linewidth}
You shouldn't plagiarize others' work. You should strive to learn in school.
\end{minipage}
&
\begin{minipage}[t]{\linewidth}
\textbf{Context:} Can I copy the work of the class before?

\textbf{Response:} You shouldn't even do that. Just do your own work and you'll be fine.
\end{minipage}
& 1 & 1 \\

\midrule

TN &
\begin{minipage}[t]{\linewidth}
You shouldn't be a misogynist.
\end{minipage}
&
\begin{minipage}[t]{\linewidth}
\textbf{Context:} I don't think that any girl should get special treatment.

\textbf{Response:} You might be confusing equal treatment with special treatment.
\end{minipage}
& 0 & 0 \\

\midrule

FN &
\begin{minipage}[t]{\linewidth}
You shouldn't take things from other people without their permission.
\end{minipage}
&
\begin{minipage}[t]{\linewidth}
\textbf{Context:} Can't you see me being hungry?

\textbf{Response:} You can always ask that person to share their food with you.
\end{minipage}
& 1 & 0 \\

\midrule

FP &
\begin{minipage}[t]{\linewidth}
You should be forgiving when others have to change their plans.
\end{minipage}
&
\begin{minipage}[t]{\linewidth}
\textbf{Context:} They made me look like a loser and I'm mad.

\textbf{Response:} I'm sure they didn't intend to make you feel like a loser. Sometimes plans change and things happen.
\end{minipage}
& 0 & 1 \\

\bottomrule
\end{tabular}
\end{table*}

\section{Case Study}

\subsection{Case Study for Full Pipeline}
We present one specific ProsocialDialog example in which the routing classifier triggers new RoT generation at a later turn. In this conversation, the initial RoT concerns public nudity in a schoolyard, capturing the broad normative issue introduced in the early turns. The following turn already moves in a more concerning direction by referring to underaged students, but the earlier RoT is still retained at that point. In Turn 3, however, the dialogue shifts more explicitly toward exposing oneself to minors, making the underlying issue both more specific and more severe than the earlier public-nudity frame. As a result, the earlier RoT is no longer sufficient to capture the updated focus of the conversation, and a new RoT is generated instead of simply reusing the previous one. The resulting prediction for this turn also matches the gold label.\footnote{Detailed case studies for each test dataset are provided in our \href{https://github.com/GitLeo1/RoTRAG-Rule-of-Thumb-Reasoning-for-Conversation-Harm-Detection-with-Retrieval-Augmented-Generation/blob/main/appendix/case-studies.md}{Github-case-studies.md}.}

\subsection{Case Study for Routing Classifier}
As shown in Figure~\ref{fig:confusion_matrix}, the routing classifier performs reliably on the validation set. Among 10,636 validation instances, it produces 7,461 true positives and 1,743 true negatives, with relatively fewer false positives and false negatives. This indicates that the classifier can effectively decide whether the previous RoT remains applicable to the current dialogue turn. The large number of true positives further suggests that the model reliably captures cases where the current response preserves the same normative core as the previous RoT.

Table~\ref{tab:classifier_examples} provides representative TP, TN, FN, and FP examples to further illustrate the classifier behavior. The correct cases show that the classifier works well when the relation between the previous RoT and the current dialogue is explicit, either by preserving the same moral principle or by clearly shifting away from it. In contrast, the error cases are mostly borderline examples. False negatives often express the same broad norm indirectly, while false positives usually involve socially reasonable responses that introduce a new condition or shift the main normative focus.
\begin{table}[t]
\centering
\small
\setlength{\tabcolsep}{4pt}
\renewcommand{\arraystretch}{1.15}
\caption{Case of the RoT routing and regeneration process.}
\label{tab:rot_routing_case}
\begin{tabular}{p{0.18\columnwidth} p{0.76\columnwidth}}
\toprule
\textbf{Field} & \textbf{Case (Turn 3)} \\
\midrule

\textbf{Previous RoT} &
You shouldn't relax naked in a public space for tanning. \\

\midrule
\textbf{Current Dialogue} &
\textbf{C}: What's wrong with exposing myself to young skin? \newline
\textbf{R}: You will be committing pedophilia if you show off your body to people under the age of 18. Do you want to be considered a pedophile just to show off your body to younger people? \\

\midrule
\textbf{Routing} &
0 (\textit{Regenerate}) \\

\midrule
\textbf{Dialogue Summary} &
Exposing oneself to minors... (\textit{abbreviated}) \\

\midrule
\textbf{Top-5 Retrieved} &
\textbf{R1}: You shouldn't get involved with underage people. \newline
\textbf{R2}: It is bad to have underage relations. \par\vspace{-0.6ex}
{\centering$\vdots$\par}\vspace{-0.4ex} \\

\midrule
\textbf{Generated RoT} &
You shouldn't expose yourself to minors. \\

\midrule
\textbf{Final Prediction} &
\textbf{Reasoning}: The current turn addresses the serious implications of exposing oneself to minors, emphasizing the legal and ethical consequences. It highlights the harmful nature of the action and frames it in a way that discourages such behavior, indicating a need for caution regarding the topic. \newline
\textbf{Label}: \textit{needs caution} \\

\midrule
\textbf{Gold Label} &
\textit{needs caution} \\

\bottomrule
\end{tabular}
\end{table}
\vspace{1em}

\section{Implications}
The results of this study suggest that multi-turn dialogue harm detection benefits from explicit normative grounding rather than relying only on a model’s internal reasoning. By retrieving and incorporating Rule of Thumb as external normative evidence, \textbf{RoTRAG} produces judgments that are more accurate, interpretable, and context-sensitive across both prosocial safety classification and multi-turn severity reasoning settings. In addition, the routing mechanism shows that stronger safety performance does not necessarily require uniformly expensive reasoning at every turn; instead, selective reasoning can provide a practical balance between effectiveness and efficiency. More broadly, these findings indicate that retrieval-grounded normative reasoning is a promising direction for building safer and more auditable conversational systems.

\section{Conclusion}
We propose \textbf{RoTRAG}, a novel framework for multi-turn dialogue harm detection that improves safety assessment through retrieval-grounded normative reasoning. Unlike prior approaches that depend mainly on parametric knowledge, prompting strategies, or internal reasoning alone, \textbf{RoTRAG} retrieves and incorporates contextually relevant Rule of Thumb as explicit normative evidence for turn-level reasoning and final prediction. Beyond the core framework, we introduce a selective routing mechanism that improves efficiency by triggering additional reasoning only when necessary, and we conduct a comprehensive evaluation across multiple dialogue safety benchmarks and backbone models. Extensive experiments show that \textbf{RoTRAG} consistently outperforms strong baselines, yielding more accurate, interpretable, and context-sensitive harm judgments. Overall, our findings highlight the value of combining retrieval and normative grounding for building safer and more reliable dialogue understanding systems.

\section*{Limitations}
Despite its strengths, this work has several limitations. First, the quality of \textbf{RoTRAG} depends on the coverage and reliability of the retrieval corpus, so missing or weakly matched Rule of Thumb examples may reduce reasoning quality. Second, Rule of Thumb is inherently simplified normative abstractions and may not fully capture cultural variation, interpersonal nuance, or conflicting social values in real-world dialogue. Third, although the encoder-based routing classifier improves efficiency, its validation accuracy of 0.8662 also indicates a trade-off, suggesting that some routing errors remain and may limit overall performance. Finally, although \textbf{RoTRAG} improves interpretability relative to purely parametric approaches, it remains an LLM-based pipeline and can still be affected by retrieval errors, generation noise, and imperfect normative reasoning.

\section*{Ethics Statement}
This work is intended to support safer and more interpretable harm detection in multi-turn dialogue by grounding model judgments in explicit Rule-of-Thumb reasoning rather than relying only on opaque parametric knowledge. At the same time, the framework raises important ethical considerations. Because the system reasons about sensitive, harmful, and socially nuanced conversations, errors may lead to over-cautious judgments, missed harms, or unfair treatment of culturally diverse expressions. In addition, retrieved Rule-of-Thumb knowledge may reflect biases, simplified moral assumptions, or incomplete normative coverage, which can affect fairness and generalizability across contexts. For this reason, RoTRAG should be used as a decision-support tool rather than as a fully autonomous moderation system, and its outputs should be interpreted with human oversight, careful dataset curation, and ongoing evaluation for bias, privacy, and potential misuse in real-world safety-sensitive applications.

\section*{Acknowledgement}
This work was supported in part by internal research funding from AI Research, Enhans AI, the National Natural Science Foundation of China (\#72474009 and \#L252400109), and the special project for discipline development at Peking University.

\bibliographystyle{ACM-Reference-Format}
\bibliography{reference}

\twocolumn

\end{document}